\documentclass[11pt,a4paper]{article}
\usepackage[nohyperref]{acl2017}
\usepackage{times}
\usepackage{latexsym}
\usepackage{amssymb}
\usepackage{amsthm}
\usepackage{color}
\usepackage{url}
\usepackage{amsmath}
\usepackage{bm}
\newcommand{\ignore}[1]{}

\usepackage{pslatex}
\usepackage{latexsym}
\usepackage{mathtools}
\usepackage{graphicx}
\usepackage{amsmath}
\usepackage{xspace}
\usepackage{subcaption}
\usepackage{tikz-dependency}
\usepackage{balance}
\usepackage{enumitem} 
\usepackage{pifont}
\usepackage{float}
\usepackage{framed} % for boxes
\usepackage{verbatim}

\DeclareSymbolFont{extraup}{U}{zavm}{m}{n}
\DeclareMathSymbol{\varheartsuit}{\mathalpha}{extraup}{86}

\newcommand{\compactparagraph}[1]{\paragraph{#1}}

\usepackage{xargs} % commandx 
\usepackage[colorinlistoftodos,prependcaption,textsize=tiny]{todonotes}
\usepackage{marginnote}

\usepackage{cleveref}
\Crefformat{equation}{(#2#1#3)}

\usepackage{microtype}
\usepackage{booktabs}
\usepackage{rotating}
\newcommandx{\todoir}[2][1=]{\todo[inline]{SR: #2}\xspace}
\newcommandx{\todoid}[2][1=]{\todo[inline]{RD: #2}\xspace}
\newcommandx{\todoiz}[2][1=]{\todo[inline]{MZ: #2}\xspace}
\newcommandx{\todoim}[2][1=]{\todo[inline]{AM: #2}\xspace}
\newcommandx{\todor}[2][1=]{\todo[linecolor=red,backgroundcolor=red!25,bordercolor=red,#1]{SR: #2}\xspace}
\newcommandx{\todod}[2][1=]{\todo[linecolor=cyan,backgroundcolor=cyan!25,bordercolor=cyan,#1]{RD: #2}\xspace}
\newcommandx{\todoz}[2][1=]{\todo[linecolor=blue,backgroundcolor=blue!10,bordercolor=blue,#1]{MZ: #2}\xspace}
\newcommandx{\todom}[2][1=]{\todo[linecolor=blue,backgroundcolor=blue!10,bordercolor=blue,#1]{AM: #2}\xspace}
\let\svthefootnote\thefootnote
\newcommand\blankfootnote[1]{%
  \let\thefootnote\relax\footnotetext{#1}%
  \let\thefootnote\svthefootnote%
}

\aclfinalcopy % Uncomment this line for the final submission
\def\aclpaperid{3478} %  Enter the acl Paper ID here

\newcommand \spades{\textsc{Spades}\xspace}

\title{Question Answering on Knowledge Bases and Text \\
using Universal Schema and Memory Networks}
\author{Rajarshi Das$^{*\spadesuit}$\quad Manzil Zaheer$^{*\heartsuit}$\quad Siva Reddy$^\clubsuit$ \and Andrew McCallum$^\spadesuit$\\
	$^\spadesuit$College of Information and Computer Sciences, University of Massachusetts Amherst\\
    $^\heartsuit$School of Computer Science, Carnegie Mellon University\\
	$^\clubsuit$School of Informatics, University of Edinburgh\\
	{\tt \{rajarshi, mccallum\}@cs.umass.edu, manzilz@cs.cmu.edu}\\
    {\tt siva.reddy@ed.ac.uk}
}
\date{}

\begin{document}
\maketitle
\begin{abstract}
Existing question answering methods infer answers either from a knowledge base or from raw text. 
While knowledge base (KB) methods are good at answering compositional questions, their performance is often affected by the incompleteness of the KB. Au contraire, %On the contrary, 
web text contains millions of facts that are absent in the KB, however in an unstructured form. {\it Universal schema} can support reasoning on the union of both structured KBs and unstructured text by aligning them in a common embedded space. In this paper we extend universal schema to natural language question answering, employing \emph{memory networks} to attend to the large body of facts in the combination of text and KB.
Our models can be trained in an end-to-end fashion on question-answer pairs.
%In this paper, we view text and KB as a unified resource for question answering. 
Evaluation results on \spades fill-in-the-blank question answering dataset show that exploiting universal schema for question answering is better than using either a KB or text alone.
This model also outperforms the current state-of-the-art by 8.5 $F_1$ points.\footnote{Code and data available in \url{https://rajarshd.github.io/TextKBQA}}
\end{abstract}

%%%%%%%%%%%%%%%%%%%%%%%%%%%%%%%%%%%%%%%%%%
\section{Introduction} % (fold)
\label{sec:introduction}
%Answering natural language questions using computers 
Question Answering (QA) has been a long-standing goal of natural language processing.
Two main paradigms evolved in solving this problem: 1)~answering  questions on a knowledge base; and 2)~answering questions using text.

Knowledge bases (KB) contains facts expressed in a fixed schema, facilitating compositional reasoning. 
These attracted research ever since the early days of computer science, e.g., BASEBALL \cite{green_jr_baseball_1961}.
This problem has matured into learning semantic parsers from parallel question and logical form pairs \cite{zelle_learning_1996,zettlemoyer_learning_2005}, to recent scaling of methods to work on very large KBs like Freebase using question and answer pairs \cite{berant_semantic_2013}. 
However, a major drawback of this paradigm is that KBs are highly incomplete \cite{dong_knowledge_2014}.
It is also an open question whether KB relational structure is expressive enough to represent world knowledge \cite{stanovsky_proposition_2014,gardner_openvocabulary_2017}

The paradigm of exploiting text for questions started in the early 1990s \cite{kupiec_murax_1993}. 
With the advent of web, access to text resources became abundant and cheap. 
Initiatives like TREC QA competitions helped popularizing this paradigm \cite{voorhees1999trec}.
%leading to the first real world Question Answering (QA) system, START \cite{katz_sentence_1997}. 
With the recent advances in deep learning and availability of large public datasets, there has been an explosion of research in a very short time \cite{rajpurkar_squad_2016,trischler_newsqa_2016,nguyen_ms_2016,wang_machine_2016,lee2016learning,xiong_dynamic_2016,seo2016query,choi2016hierarchical}.
Still, text representation is unstructured and does not allow the compositional reasoning which structured KB supports.

%%%%%%%%%%%%%%%%%%%%%%%%%%%%%%%%%%%%%%%%%%%
%\begin{figure*}
%\vspace{-4mm}
%\begin{minipage}[!b][][t]{.62\textwidth}
%\centering
%\includegraphics[width=0.9\linewidth]{images/USchema_QA_Attention.pdf}
%\captionof{figure}{ Memory network attending the facts in the universal schema (matrix on the left). The color gradients denote the attention weight on each fact}
%\label{fig:model}
%\vspace{-2mm}
%\end{minipage}\hfill
%\begin{minipage}[!b][][t]{.32\textwidth}
%\centering
%\includegraphics[width=\linewidth]{images/sweep1.png}
%\captionof{figure}{Performance on varying the number of available KB facts during test time. \textsc{UniSchema} model consistently outperforms \textsc{OnlyKB}}
%\label{fig:kb_coverage}
%\end{minipage}
%\vspace{-4mm}
%\end{figure*}
%
\begin{figure*}
\small
\centering
	\includegraphics[width=1.6\columnwidth]{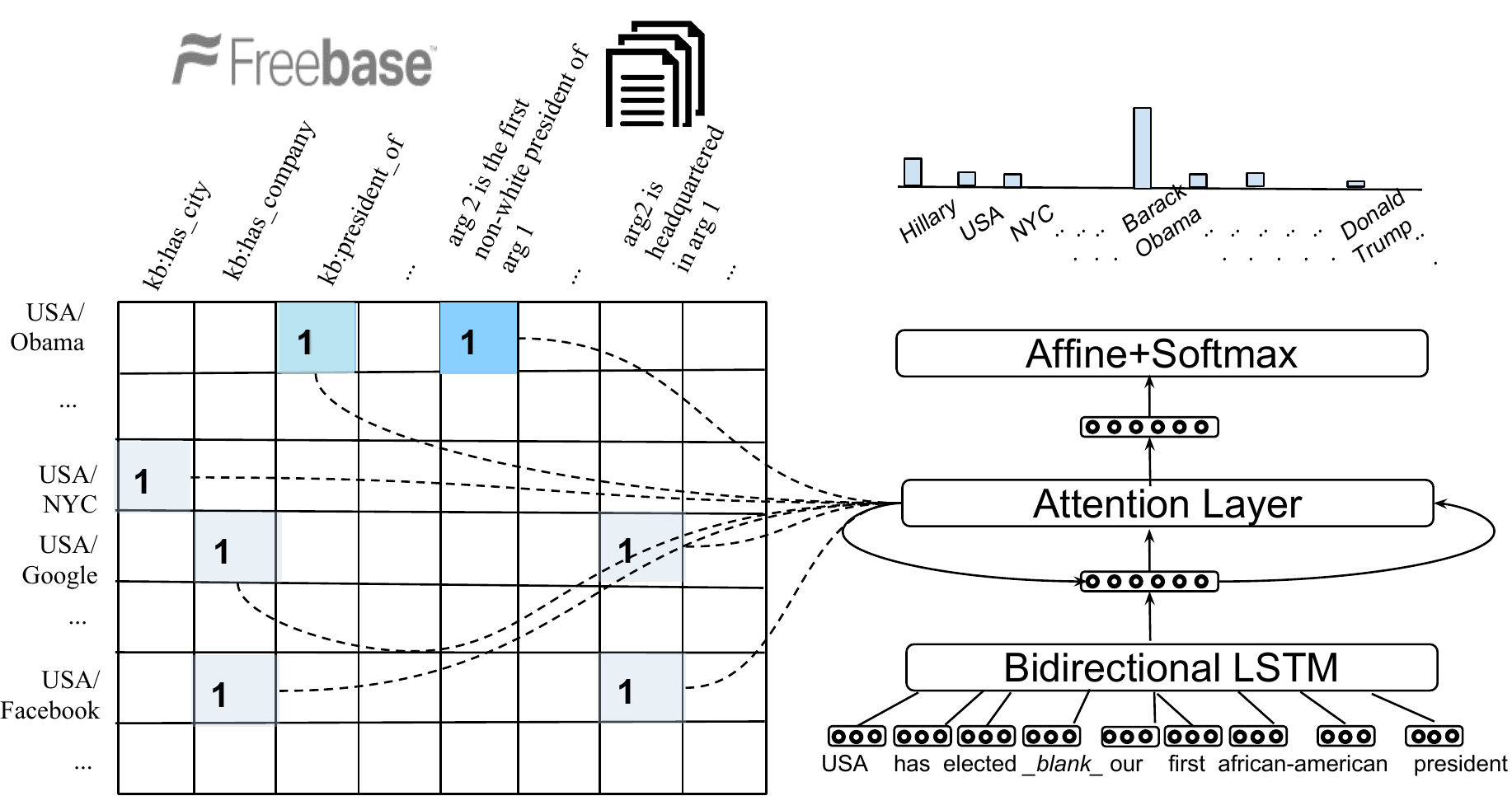}
\caption{Memory network attending the facts in the universal schema (matrix on the left). The color gradients denote the attention weight on each fact.}
\label{fig:model}
\end{figure*} 
%%%%%%%%%%%%%%%%%%%%%%%%%%%%%%%%%%%%%%%%%%%

An important but under-explored QA paradigm is where KB and text are exploited together \cite{ferrucci_building_2010}.
Such combination is attractive because text contains millions of facts not present in KB, and a KB's generative capacity represents infinite number of facts that are never seen in text.
However QA inference on this combination is challenging due to the structural non-uniformity of KB and text.
%KBs are incomplete \cite{min13} and text can express many relations which are not present in KBs (e.g. `long time friend'). 
\emph{Distant supervision} methods \cite{bunescu07,mintz_ds,Riedel:2010,Yao:2010,zeng15} address this problem partially by means of aligning text patterns with KB. 
But the rich and ambiguous nature of language allows a fact to be expressed in many different forms which these models fail to capture. 
\emph{Universal schema} \cite{USchema:13} avoids the alignment problem by jointly embedding KB facts and text into a uniform structured representation, allowing interleaved propagation of information.
\Cref{fig:model} shows a universal schema matrix which has pairs of entities as rows, and Freebase and textual relations in columns. Although universal schema has been extensively used for relation extraction, this paper shows its applicability to QA. Consider the question \textsl{USA has elected \_blank\_, our first african-american president} with its answer \textsl{Barack Obama}. While Freebase has a predicate for representing presidents of USA, it does not have one for `african-american' presidents.
Whereas in text, we find many sentences describing the presidency of Barack Obama and his ethnicity at the same time.
Exploiting both KB and text makes it relatively easy to answer this question than relying on only one of these sources.

Memory networks (MemNN; \citealt{MemNN}) are a class of neural models which have an external memory component for encoding short and long term context. 
In this work, we define the memory components as observed cells of the universal schema matrix, and train an end-to-end QA model on question-answer pairs.

% Specifically we represent Universal Schema as key-value pairs in the memory (\Cref{sec:}), keys formed by subjects of relations and relations themseleved, the objects as relations.}
The contributions of the paper are as follows (a)~We show that universal schema representation is a better knowledge source for QA than either KB or text alone, (b) On the SPADES dataset \cite{bisk_evaluating_2016}, containing real world fill-in-the-blank questions, we outperform state-of-the-art semantic parsing baseline, with 8.5~$F_1$ points. (c) Our analysis shows how individual data sources help fill the weakness of the other, thereby improving overall performance.

\section{Background}

\compactparagraph{Problem Definition}  Given a question $q$ with words $w_1, w_2, \ldots, w_n$, where these words contain one $\_blank\_$ and at least one entity, our goal is to fill in this $\_blank\_$ with an answer entity $q_a$ using a knowledge base $\mathcal{K}$ and text $\mathcal{T}$.		
Few example question answer pairs are shown in Table~\ref{tbl:sentences}.

\compactparagraph{Universal Schema}
Traditionally universal schema is used for relation extraction in the context of knowledge base population.
Rows in the schema are formed by entity pairs (e.g. USA, NYC), and columns represent the relation between them.
A relation can either be a KB relation, or it could be a pattern of text that exist between these two entities in a large corpus. The embeddings of entities and relation types are learned by low-rank matrix factorization techniques.
\newcite{USchema:13} treat textual patterns as static symbols, whereas recent work by \newcite{pat:15} replaces them with distributed representation of sentences obtained by a RNN.
Using distributed representation allows reasoning on sentences that are similar in meaning but different on the surface form. 
We too use this variant to encode our textual relations.

%%%%%%%%%%%%%%%%%%%%%%%%%%%%%%%%%%%%%%%%%%%
%\begin{table}[t]
%\vspace{-6mm}
%\small
%\centering
%\begin{tabular}{p{0.7\columnwidth}l}
%\toprule
%\multicolumn{1}{c}{\bf Question} & \multicolumn{1}{c}{\bf Answer} \\
%\midrule
%NBC\_Universal is owned by \_blank\_ & GE \\
%Pixar is based in \_blank\_, CA & Emeryville \\
%\_blank\_ is the biggest city in the Benelux &  Brussels \\
%\bottomrule
%\end{tabular}
%\vspace{-2mm}
%\caption{\label{tbl:examples} Questions with blanks and their answers}
%\vspace{-6mm}
%\end{table}

%%%%%%%%%%%%%%%%%%%%%%%%%%%%%%%%%%%%%%%%%%

\compactparagraph{Memory Networks}
MemNNs are neural attention models with external and differentiable memory. MemNNs decouple the memory component from the network thereby allowing it store external information.
Previously, these have been successfully applied to question answering on KB where the memory is filled with distributed representation of KB triples \cite{simple_qa}, or for reading comprehension \cite{E2EMemNN,goldilocks}, where the memory consists of distributed representation of sentences in the comprehension.
Recently, key-value MemNN are introduced \cite{KVMemNN} where each memory slot consists of a key and value.
The attention weight is computed only by comparing the question with the key memory, whereas the value is used to compute the contextual representation to predict the answer. We use this variant of MemNN for our model. 
\newcite{KVMemNN}, in their experiments, store either KB triples or sentences as memories but they do not explicitly model multiple memories containing distinct data sources like we do.

\section{Model}
Our model is a MemNN with universal schema as its memory.
\Cref{fig:model} shows the model architecture.

\paragraph{Memory:}
Our memory $\mathcal{M}$ comprise of both KB and textual triples from universal schema.
Each memory cell is in the form of key-value pair.
Let $(\mathrm{s,r,o}) \in \mathcal{K}$ represent a KB triple.
We represent this fact with distributed key $\mathbf{k} \in \mathbb{R}^{2d}$  formed by concatenating the embeddings $\mathbf{s} \in \mathbb{R}^{d}$ and $\mathbf{r} \in \mathbb{R}^{d}$ of subject entity $s$ and relation $r$ respectively.
The embedding $\mathbf{o} \in \mathbb{R}^{d}$ of object entity $o$ is treated as its value $\mathbf{v}$.

Let $(\mathrm{s,~[w_1,\ldots,arg_1,\ldots, arg_2,w_n],~o}) \in \mathcal{T}$ represent a textual fact, where $\mathrm{arg_1}$ and $\mathrm{arg_2}$ correspond to the positions of the entities `$\mathrm{s}$' and `$\mathrm{o}$'.
We represent the key as the sequence formed by replacing $\mathrm{arg_1}$ with `$s$' and $arg_2$ with a special `$\_\mathrm{blank}\_$' token, i.e., $k$ = $[\mathrm{w_1},~\ldots,\mathrm{s},~\ldots,~\_\mathrm{blank}\_,~\mathrm{w_n}]$ and value as just the entity `$\mathrm{o}$'.
We convert $k$ to a distributed representation using a bidirectional LSTM \cite{lstm,bilstm}, where $\mathbf{k} \in \mathbb{R}^{2d}$ is formed by concatenating the last states of forward and backward LSTM, i.e., $\mathbf{k}~=~ \left[\overrightarrow{\mathrm{LSTM}}(\mathrm{k});\overleftarrow{\mathrm{LSTM}}(\mathrm{k})\right]$. The value $\mathbf{v}$ is the embedding of the object entity $o$.
Projecting both KB and textual facts to $\mathbb{R}^{2d}$ offers a unified view of the knowledge to reason upon. 
In \Cref{fig:model}, each cell in the matrix represents a memory containing the distributed representation of its key and value.

\paragraph{Question Encoder:} A bidirectional LSTM is also used to encode the input question $q$ to a distributed representation  $\mathbf{q} \in \mathbb{R}^{2d}$ similar to the key encoding step above.

\paragraph{Attention over cells:} 
We compute attention weight of a memory cell by taking the dot product of its key $\mathbf{k}$ with a contextual vector $\mathbf{c}$ which encodes most important context in the current iteration.
In the first iteration, the contextual vector is the question itself.
We only consider the memory cells that contain at least one entity in the question.
For example, for the input question in \Cref{fig:model}, we only consider memory cells containing~USA.
Using the attention weights and values of memory cells, we compute the context vector $\mathbf{c_t}$ for the next iteration $t$ as follows:
\vspace{-5pt}
\begin{equation*}
\vspace{-5pt}
\mathbf{c_t} = \mathbf{W_t} \left( \mathbf{c_{t-1}} + \mathbf{W_p}\sum_{(k,v) \in \mathcal{M}} (\mathbf{c_{t-1}} \cdot \mathbf{k}) \mathbf{v} \right)
\end{equation*}
where $\mathbf{c_0}$ is initialized with question embedding 
$\mathbf{q}$, $\mathbf{W_p}$ is a projection matrix, and 
$\mathbf{W_t}$ represents the weight matrix which considers the 
context in previous hop and the values in the current iteration based on their importance (attention weight).
This multi-iterative context selection allows multi-hop reasoning without explicitly requiring a symbolic query representation.

\paragraph{Answer Entity Selection:} The final contextual vector $c_t$ is used to select the answer entity $q_a$ (among all 1.8M entities in the dataset) which has the highest inner product with it.
% \vspace{-6pt}
% \begin{equation*}
% \vspace{-6pt}
% \mathbf{q_a} = \underset{e}{\argmax} \;\;\mathbf{e} \cdot \mathbf{c_t}
% \end{equation*}
% Finally, we minimize the cross-entropy loss to train the complete MemNN in an end-to-end fashion.

%%%%%%%%%%%%%%%%%%%%%%%%%%%%%%%%%%%%%%%%%%%

\section{Experiments} % (fold)
\label{sec:experiments}
% We describe our evaluation datasets, implementation details, baseline models and results.
\subsection{Evaluation Dataset}
\label{sub:dataset}
We use Freebase \cite{fb} as our KB, and ClueWeb \cite{clue_web1} as our text source to build universal schema.
For evaluation, literature offers two options: 1) datasets for text-based question answering tasks such as answer sentence selection and reading comprehension; and 2) datasets for KB question answering.

Although the text-based question answering datasets are large in size, e.g., SQuAD \cite{rajpurkar_squad_2016} has over 100k questions, answers to these are often not entities but rather sentences which are not the focus of our work. Moreover these texts may not contain Freebase entities at all, making these skewed heavily towards text. 
Coming to the alternative option, WebQuestions \cite{berant_semantic_2013} is widely used for QA on Freebase.
This dataset is curated such that all questions can be answered on Freebase alone. 
But since our goal is to explore the impact of universal schema, testing on a dataset completely answerable on a KB is not ideal.
WikiMovies dataset \cite{KVMemNN} also has similar properties.
%since they have carefully curated a KB which can answer all questions in the dataset.
\newcite{gardner_openvocabulary_2017} created a dataset with motivations similar to ours, however this is not publicly released during the submission time.

%%%%%%%%%%%%%%%%%%%%%%%%%%%%%%%%%%%%%%%%%%

\begin{table}
\vspace{2mm}
\small
\centering
\begin{tabular*}{\columnwidth}{l@{\extracolsep{1em}}@{\extracolsep{\fill}}c c@{\extracolsep{2em}}c@{\extracolsep{1em}}c c@{\extracolsep{2em}}c@{\extracolsep{1em}}c @{\extracolsep{\fill}}}
\toprule
\bf Model & \bf Dev. F$_1$ & \bf Test F$_1$\\
\midrule 
\newcite{bisk_evaluating_2016} & 32.7 & 31.4 \\
\textsc{OnlyKB}  & 39.1 & 38.5 \\
\textsc{OnlyText} & 25.3 & 26.6 \\
\textsc{Ensemble.} & 39.4 & 38.6 \\
\textsc{UniSchema} & \textbf{41.1} & \textbf{39.9} \\ 
\bottomrule
\end{tabular*}
\vspace{-1mm}
\caption{QA results on \spades.}
\label{tbl:res}
\vspace{-3mm}
\end{table}

%%%%%%%%%%%%%%%%%%%%%%%%%%%%%%%%%%%%%%%%%%

Instead, we use \textsc{Spades} \cite{bisk_evaluating_2016} as our evaluation data which contains fill-in-the-blank cloze-styled questions created from ClueWeb.
This dataset is ideal to test our hypothesis for following reasons: 1) it is large with 93K sentences and 1.8M entities; and 2) since these are collected from Web, most sentences are natural.
A limitation of this dataset is that it contains only the sentences that have entities connected by at least one relation in Freebase, making it skewed towards Freebase as we will see (\S~\ref{sub:results}).
We use the standard train, dev and test splits for our experiments. For text part of universal schema, we use the sentences present in the training set. 

%\subsection{Implementation Details}
%hyperparameters are optimized on the development set.
% We found that the network didn't train well when we naively combined the KB and textual facts. We found that the network always converged to a  local minimum \todo{I would not say it is a local minima, because training loss keeps decreasing} closer  to the performance of a network which had only textual facts as evidences. We tried an approach similar to batch normalization \cite{batch_norm}, in which for each minibatch, we normalize the mean and variance of the textual facts and then scale and shift to match the mean and variance of the KB memory. Empirically, this stabilized training and gave a boost in the final performance.
%The performance is reported using the accuracy of the top hit (single answer) over all possible answers (all entities), i.e. the hits@1 metric measured in percent.
%We used the Adam optimizer for training with an initial learning rate
%of~0.001, two momentum parameters [0.99, 0.999] and a batch size of 32. \todo{i don't this is important and taking space}

\subsection{Models}
We evaluate the following models to measure the impact of different knowledge sources for QA. 

\paragraph{\textsc{OnlyKB:}} In this model, MemNN memory contains only the facts from KB.
For each KB triple $(e_1,r,e_2)$, we have two memory slots, one for $(e_1,r,e_2)$ and the other for its inverse $(e_2,r^i,e_1)$.

\paragraph{\textsc{OnlyTEXT:}}\spades contains sentences with blanks. 
We replace the blank tokens with the answer entities to create textual facts from the training set.
Using every pair of entities, we create a memory cell similar to as in universal schema.

\paragraph{\textsc{Ensemble}} This is an ensemble of the above two models. We use a linear model that combines the scores from, and use an ensemble to combine the evidences from individual models.

\paragraph{\textsc{UniSchema}} This is our main model with universal schema as its memory, i.e., it contains memory slots corresponding to both KB and textual facts.

\subsection{Implementation Details}

The dimensions of word, entity and relation 
embeddings, and LSTM states were set to $d= $50. 
The word and entity embeddings were initialized with word2vec \cite{word2vec} trained on 7.5 million ClueWeb sentences containing entities in Freebase subset of \spades.
The network weights were initialized using Xavier initialization \cite{glorot}.
We considered up to a maximum of 5k KB facts and 2.5k textual facts for a question. 
We used Adam \cite{adam} with the default hyperparameters (learning rate=1e-3, $\beta_1$=0.9, $\beta_2$=0.999, $\epsilon$=1e-8) for optimization. To overcome exploding gradients, we restricted the magnitude of the $\ell_2$ norm of the gradient to 5. The batch size during training was set to 32.

To train the \textsc{UniSchema} model, we initialized the parameters from a trained \textsc{OnlyKB} model.
We found that this is crucial in making the \textsc{UniSchema} to work.
% We also found that the network doesn't train well when we naively combine the KB and textual facts. 
Another caveat is the need to employ a trick similar to \emph{batch normalization} \cite{batch_norm}. 
For each minibatch, we normalize the mean and variance of the textual facts and then scale and shift to match the mean and variance of the KB memory facts. Empirically, this stabilized the training and gave a boost in the final performance.

%%%%%%%%%%%%%%%%%%%%%%%%%%%%%%%%%%%%%%%%%%

\begin{table}[t]
\vspace{2mm}
\small
\centering
\begin{tabular}{p{0.70\columnwidth}l}
\toprule
\multicolumn{1}{c}{\bf Question} & \multicolumn{1}{l}{\bf Answer} \\
\midrule
1. USA have elected \_blank\_, our first african-american president.  & Obama \\
2. Angelina has reportedly been threatening to leave \_blank\_. & Brad\_Pitt \\
3. Spanish is more often a second and weaker language among many \_blank\_. & Latinos \\
4. \_blank\_  is the third largest city in the United\_States. & Chicago \\
5. \_blank\_ was Belshazzar 's father. & Nabonidus \\
\bottomrule
\end{tabular}
\caption{A few questions on which \textsc{OnlyKB} fails to answer but \textsc{UniSchema} succeeds.}
\label{tbl:sentences}
\vspace{-3mm}
\end{table}

%%%%%%%%%%%%%%%%%%%%%%%%%%%%%%%%%%%%%%%%%%

\subsection{Results and Discussions}
\label{sub:results}
\Cref{tbl:res} shows the main results on \spades.
\textsc{UniSchema} outperforms all our models validating our hypothesis that exploiting universal schema for QA is better than using either KB or text alone.
Despite \spades creation process being friendly to Freebase, exploiting text still provides a significant improvement.
\Cref{tbl:sentences} shows some of the questions which \textsc{UniSchema} answered but \textsc{OnlyKB} failed.
These can be broadly classified into (a) relations that are not expressed in Freebase (e.g., african-american presidents in sentence~1); (b) intentional facts since curated databases only represent concrete facts rather than intentions (e.g., threating to leave in sentence~2); (c) comparative predicates like first, second, largest, smallest (e.g.,~sentences~3 and~4); and (d) providing additional type constraints (e.g., in sentence~5, Freebase does not have a special relation for \textsl{father}.
It can be expressed using the relation \textsl{parent} along with the type constraint that the answer is of gender \textsl{male}).

We have also anlalyzed the nature of \textsc{UniSchema} attention.
In 58.7\% of the cases the attention tends to prefer KB facts over text.
This is as expected since KBs facts are concrete and accurate than text.
%, and the network figures out it is a more reliable source than text.
In 34.8\% of cases, the memory prefers to attend text even if the fact is already present in the KB.
For the rest (6.5\%), the memory distributes attention weight evenly, indicating for some questions, part of the evidence comes from text and part of it from KB.
\Cref{tbl:resContrast} gives a more detailed quantitative analysis of the three models in comparison with each other.

\begin{figure}
\vspace{-1mm}
\small
\centering
\includegraphics[scale=0.54,trim={0 1mm 0 0},clip]{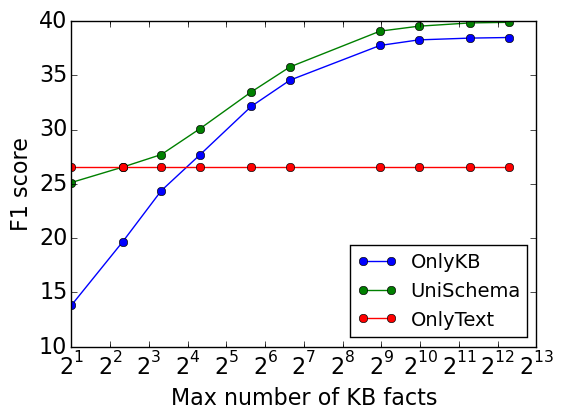}
\vspace{-1mm}
\caption{Performance on varying the number of available KB facts during test time. \textsc{UniSchema} model consistently outperforms \textsc{OnlyKB}}
\label{fig:kb_coverage}
\vspace{-2mm}
\end{figure}

To see how reliable is \textsc{UniSchema}, we gradually increased the coverage of KB by allowing only a fixed number of randomly chosen KB facts for each entity.
As Figure~\ref{fig:kb_coverage} shows, when the KB coverage is less than~16 facts per entity, \textsc{UniSchema} outperforms \textsc{OnlyKB} by a wide-margin indicating \textsc{UniSchema} is robust even in resource-scarce scenario, whereas \textsc{OnlyKB} is very sensitive to the coverage. \textsc{UniSchema} also outperforms \textsc{Ensemble} showing joint modeling is superior to ensemble on the individual models.
We also achieve the state-of-the-art with~8.5 $F_1$ points difference.
\citeauthor{bisk_evaluating_2016} use graph matching techniques to convert natural language to Freebase queries whereas even without an explicit query representation, we outperform them.

%%%%%%%%%%%%%%%%%%%%%%%%%%%%%%%%%%%%%%%%%%

% Note that as the SPADES dataset is automatically generated it is 
% extremely noisy, and in some cases our methods retrieves the correct 
% answers which are incorrect in the dataset!

%%%%%%%%%%%%%%%%%%%%%%%%%%%%%%%%%%%%%%%%%%

\begin{table}
\small
\centering
\begin{tabular*}{\columnwidth}{l@{\extracolsep{1em}}@{\extracolsep{\fill}}c c@{\extracolsep{2em}}c@{\extracolsep{1em}}c c@{\extracolsep{2em}}c@{\extracolsep{1em}}c @{\extracolsep{\fill}}}
\toprule
\bf Model & \bf Dev. F$_1$\\
\midrule 
\textsc{OnlyKB} correct 		& 39.1 \\
\textsc{OnlyText} correct 		& 25.3 \\
\textsc{UniSchema} correct 			& 41.1 \\
\textsc{OnlyKB} or \textsc{OnlyText} got it correct &	45.9 \\
\\
Both \textsc{OnlyKB} and \textsc{OnlyText} got it correct 	& 18.5 \\
\textsc{OnlyKB} got it correct and \textsc{OnlyText} did not 	& 20.6 \\
\textsc{OnlyText} got it correct and \textsc{OnlyKB} did not 	& 6.80 \\
\\
Both \textsc{UniSchema} and \textsc{OnlyKB} got it correct 	& 34.6 \\
\textsc{UniSchema} got it correct and \textsc{OnlyKB} did not 	& 6.42 \\
\textsc{OnlyKB} got it correct and \textsc{UniSchema} did not 	& 4.47 \\
\\
Both \textsc{UniSchema} and \textsc{OnlyText} got it correct 	& 19.2 \\
\textsc{UniSchema} got it correct and \textsc{OnlyText} did not 	& 21.9 \\
\textsc{OnlyText} got it correct and \textsc{UniSchema} did not 	& 6.09 \\

%kbtext got correct which either of \textsc{OnlyKB} or text got it correct 	& 36.7 \\
%kbtext got correct which neither of \textsc{OnlyKB} nor text got it correct & 4.33 \\
\bottomrule
\end{tabular*}
\caption{Detailed results on \spades.}
\label{tbl:resContrast}
\vspace{-2mm}
\end{table}

%%%%%%%%%%%%%%%%%%%%%%%%%%%%%%%%%%%%%%%%%%

%%%%%%%%%%%%%%%%%%%%%%%%%%%%%%%%%%%%%%%%%%

%\begin{table}
%\small
%\begin{tabular}{p{\columnwidth}}
%\toprule

%\bottomrule
%\end{tabular}
%\caption{Attention on unified memory}
%\end{table}

%%%%%%%%%%%%%%%%%%%%%%%%%%%%%%%%%%%%%%%%%%%
\section{Related Work}
% 
% START Omnibase: Uniform Access to Heterogeneous Data for Question Answering
% Why challenging
%% Training data format
%% Compositional reasoning on text -- traditional way is to convert text to IE triples. Memory networks are good here. Universal Schema. Hill et al work.
%% 

A majority of the QA literature that focused on exploiting KB and text either improves the inference on the KB using text based features \cite{krishnamurthy_weakly_2012,reddy_largescale_2014,joshi_knowledge_2014,yao_information_2014,yih_semantic_2015,neelakantan15,gu15,xu_question_2016,choi_scalable_2015,savenkov_when_2016} or improves the inference on text using KB \cite{sun_open_2015}.

Limited work exists on exploiting text and KB jointly for question answering.
\newcite{gardner_openvocabulary_2017} is the closest to ours who generate a open-vocabulary logical form and rank candidate answers by how likely they occur with this logical form both in Freebase and text.
Our models are trained on a weaker supervision signal without requiring the annotation of the logical forms.
%The main difference is they require a CCG parser trained on a manual treebank to parse sentences to logical forms whereas our end-to-end trained models  also learn a latent representation for sentences.

A few QA methods infer on curated databases combined with OpenIE triples \cite{fader_open_2014,yahya_relationship_2016,xu_hybrid_2016}.
Our work differs from them in two ways: 1) we do not need an explicit database query to retrieve the answers \cite{neelakantan2015neural,andreas_learning_2016}; and 2) our text-based facts retain complete sentential context unlike the OpenIE triples \cite{banko_open_2007,carlson_toward_2010}.

% Our work also intersects with vast literature on relation extraction and knowledge-base population using text \cite{toutanova_representing_2015,pat:15,pat:16}. Recently compositional models \cite{neelakantan15,gu15,toutanova16,das17} have been proposed which reasons on paths between entity pairs.
% and especially with universal schema \cite{USchema:13}.
% Ours is the first work to use Universal Schema for QA.\todod{may be dont need this?}
% While earlier works use matrix factorization for inference on Universal Schema, we use Memory Networks.
%While in the previous works, the number of columns in Universal Schema requires combinatorial memory for each sentence, we require only linear number of columns since we store each sentence as key-value pairs, where key is formed by removing one of the entities in the sentence, and the value is the entity removed.\todor{Not sure if we want to say this. Looks like waste of space :) and probably confusing since Figure does not reflect this.}

% Compositional

% Methods that use both KB and Text in a pipeline fashion
% Kun, Yahya, Sun, Joshi
% Closest to our method is DeepQA.

\section{Conclusions}
In this work, we showed universal schema is a promising knowledge source for QA than using KB or text alone.
Our results conclude though KB is preferred over text when the KB contains the fact of interest, a large portion of queries still attend to text indicating the amalgam of both text and KB is superior than KB alone.

\section*{Acknowledgments}
We sincerely thank Luke Vilnis for helpful insights. This work was supported in part by the Center for Intelligent Information Retrieval and in part by DARPA under agreement number FA8750-13-2-0020. The U.S. Government is authorized to reproduce and distribute reprints for Governmental purposes notwithstanding any copyright notation thereon. Any opinions, findings and conclusions or recommendations expressed in this material are those of the authors and do not necessarily reflect those of the sponsor.

% section experiments (end)

\bibliography{acl2017}
\bibliographystyle{acl_natbib}

\end{document}